\documentclass[letterpaper, 10 pt, conference]{ieeeconf}  
\IEEEoverridecommandlockouts                                  \usepackage{amsmath}
\usepackage{amssymb}
\usepackage{mathtools}
\usepackage{graphicx}
\usepackage{float}
\usepackage{array}
\usepackage{tikz}
\usepackage{listings}
\usepackage{hyperref}
\usepackage{graphicx}
\usepackage{epstopdf}
\usepackage{diagbox}
\usepackage{cite}
\usepackage{dsfont}
\usepackage{pdfpages}
\usepackage{mathtools,etoolbox}
\usepackage[none]{hyphenat}
\usepackage[utf8]{inputenc}
\usepackage[english]{babel}

\usepackage{amsthm}
\usepackage{xfrac}
\usepackage{nicefrac}
\usepackage{booktabs}
\usepackage[switch, pagewise]{lineno}
\usepackage{nicefrac}
\usepackage{flushend}
\usepackage{xcolor}
\usepackage{todonotes}

\long\def\invis#1{}

\usepackage{multicol}
\usepackage{algorithm}
\usepackage{amsmath}
\usepackage{algpseudocode}
\usepackage{xcolor}
\usepackage{graphicx}
\usepackage{gensymb}
\usepackage{cite}
\usepackage{subcaption}
\newcommand{\sys}{\textit{\textbf{DREAM}}}

\title{\LARGE \bf \sys : Domain-aware Reasoning for Efficient Autonomous Underwater Monitoring
}

\author{
Zhenqi Wu$^{1}$*, Abhinav Modi$^{2}$*, Angelos Mavrogiannis$^{2}$, Kaustubh Joshi$^{2}$, \\
Nikhil Chopra$^{2}$, Yiannis Aloimonos$^{2}$, Nare Karapetyan$^{3}$, Ioannis Rekleitis$^{4}$, Xiaomin Lin$^{1,\dagger}$,
\thanks{This work was supported by USDA NIFA sustainable agriculture system program under award
number 20206801231805.}
\thanks{* Equal Contributors, $\dagger$ Corresponding Author}
\thanks{$^{1}$Electrical Engineering, University of South Florida, Tampa, FL, 33620, USA. Emails:\texttt{\{zhenqi, xlin2\}@usf.edu}}
\thanks{$^{2}$Maryland Robotics Center (MRC), University of Maryland, College Park, MD 20742, USA. Emails: \texttt{\{abhi1625, angelosm, kjoshi, nchopra, jyaloimo\}@umd.edu}.}
\thanks{$^{3}$Woods Hole Oceanographic Institution (WHOI), Woods Hole, MA, 02543, USA. Emails: {\tt\small nare@whoi.edu}}
\thanks{$^{4}$Mechanical Engineering, University of Delaware, Newark, DE 19716, USA. Emails: {\tt\small yiannisr@udel.edu}}
}

\begin{document}

\maketitle
\thispagestyle{empty}
\pagestyle{empty}


\begin{abstract}
The ocean is warming and acidifying, increasing the risk of mass mortality events for temperature-sensitive shellfish such as oysters. This motivates the development of long-term monitoring systems. However, human labor is costly and long-duration underwater work is highly hazardous, thus favoring robotic solutions as a safer and more efficient option. To enable underwater robots to make real-time, environment-aware decisions without human intervention, we must equip them with an intelligent “brain”. This highlights the need for persistent, wide-area, and low-cost benthic monitoring. To this end, we present \sys, a Vision Language Model (VLM)-guided autonomy framework for long-term underwater exploration and habitat monitoring.
\sys~couples (i) a reasoning-augmented prompt that guides VLM planning with (ii) an occupancy map providing memory and overview, and (iii) a low-level controller to realize actions.
The results show that our framework is highly efficient in finding and exploring target objects (e.g., oysters, shipwreck) without prior location information. In the oyster-monitoring task, our framework takes 31.5\% less time than previous baseline with the same amount of oysters. Compared to the vanilla VLM, it uses 23\% fewer steps while covering 8.88\% more oysters. In shipwreck scenes, our framework successfully explores and maps the wreck without collisions, requiring 27.5\% fewer steps than the vanilla model and achieving 100\% coverage, while the vanilla model achieves 60.23\% average coverage in our shipwreck environments. 
\end{abstract}
\section{Introduction}
Oceans are the lifeline of our planet, nurturing entire ecosystems\cite{ward2022safeguarding} that are essential to human survival and controlling the global climate of the earth\cite{bigg2003role}. Despite their vast size, a significant portion of the oceans has not yet been explored or mapped, hindering our ability to safeguard vital ecosystems and anticipate their responses to global environmental shifts~\cite{mckenzie2020global}. Direct human exploration can often be too risky. For example, monitoring a coral reef might require maneuvering through confined areas or environments that often lie beyond the reach of human divers due to additional adverse conditions such as extreme pressure, limited visibility, and the physical risks involved.

Deploying Remotely Operated Vehicles (ROVs) offers a safer and more practical solution but can be challenging in uncharted underwater environments and inefficient for long-term monitoring~\cite{li2023bioinspired}. ROV operators often need to collaborate closely with ocean scientists, but differences in background and expertise can complicate communication. Furthermore, the survival of vital ecosystems requires long-term continuous monitoring, which can make the entire operation prohibitively expensive due to the costs associated with logistics, specialized personnel, and equipment~\cite{mclean2020enhancing}. Vision-based monitoring methods have shown promise in specific domains, with applications in surveying coral reefs~\cite{manderson2018vision}, shipwrecks~\cite{karapetyan2021human}, and oysters~\cite{lin2024uivnav}. However, these methods are inherently imitation learning-based and have no feedback from the outer world. Recent systems such as OceanPlan~\cite{yang2024oceanplan}, OceanChat~\cite{yang2023oceanchat}, and AquaChat++~\cite{saad2025aquachat++} advance natural language piloting, and Xu et al.~\cite{10564009,11247231,xu2025cockycooperatefimrlbased} contribute multiple works on multi-AUV decision making, coordination, and adaptive control, but still lack robust reasoning, adaptability, and persistent spatial memory. 
\begin{figure}[t]
    \centering
    {\includegraphics[width=\columnwidth]{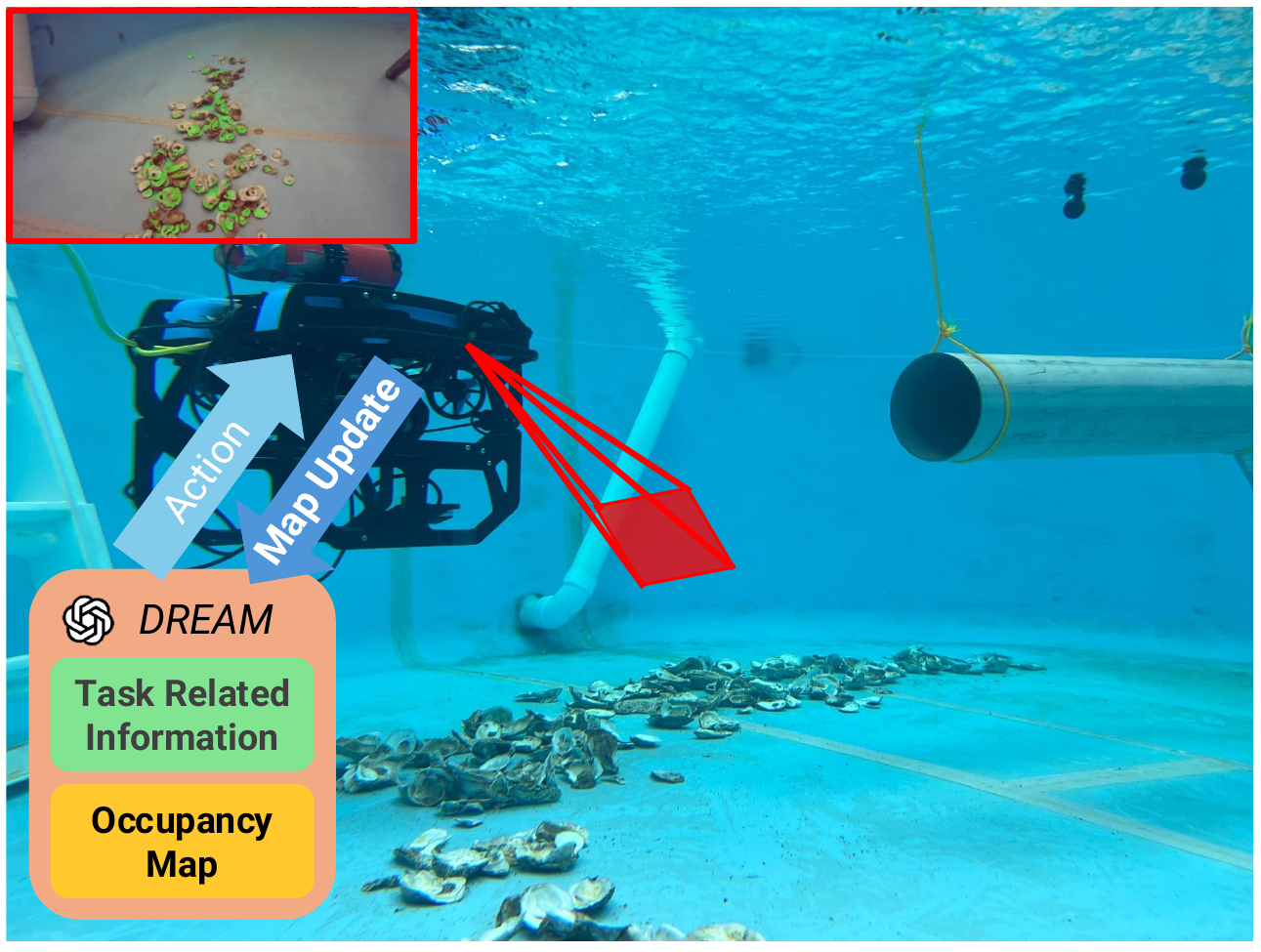}}
    \vspace{-3mm}
    \caption{Example of \sys~ deployed in the real world on a BlueROV surveying an oyster reef in a pool. The top left image shows a sample observation from the robot's camera.}
    \label{fig:beauty_shot}
    \vspace{-4mm}
\end{figure}
\begin{figure*}[ht!]
\vspace{3mm}
\includegraphics[width=0.95\textwidth]{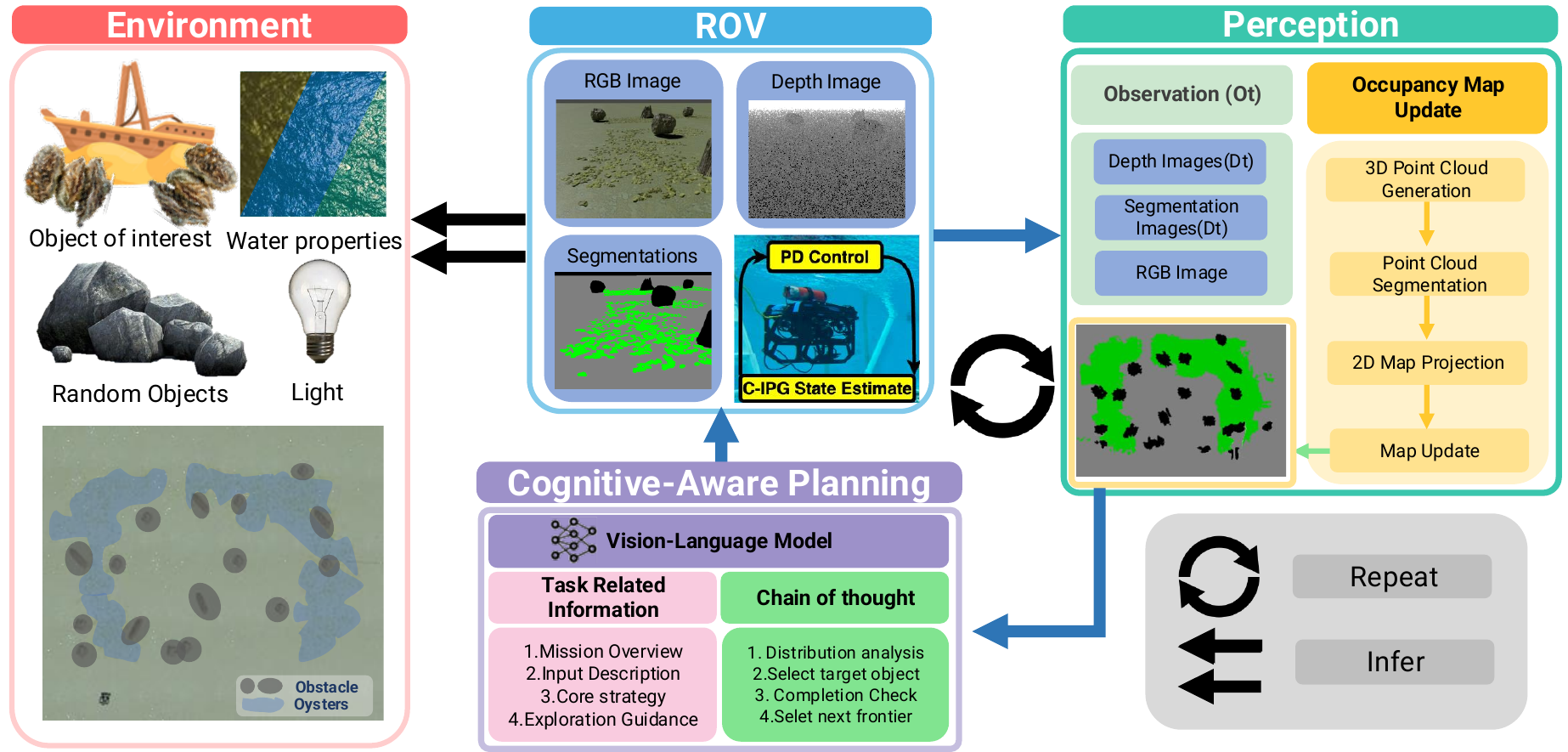}
\centering
\caption{An overview of the DREAM framework. The environment provides multimodal inputs (RGB, depth, and segmentation) captured by the ROV. These observations are fused in the Perception module to build and update occupancy maps. Cognitive-Aware Planning leverages a vision–language model with chain-of-thought reasoning to guide frontier selection, mission planning, and persistent monitoring of underwater objects of interest. Robotic control then executes movement commands, closing the loop for continual exploration and monitoring.}
\vspace{-5mm}
\label{fig:overview}
\end{figure*}
To address these limitations, we propose a new VLM-based underwater monitoring System, \textbf{\textit{D}}omain-aware \textbf{\textit{R}}easoning for \textbf{\textit{E}}fficient \textbf{\textit{A}}utonomous 
Underwater \textbf{\textit{M}}onitoring (\sys). The system leverages human diver knowledge, oceanic domain knowledge, together with real-time visual input from an onboard camera to generate adaptive monitoring policies that underwater robots can execute. The main contributions of this work are:
\begin{enumerate}
\item A 3-layer architecture of perception, cognitive-aware planning, and control, that enables end-to-end monitoring — from perception to high-level reasoning and finally to low-level adaptive execution.
\item A VLM-based framework that integrates domain knowledge with sequential reasoning (via Chain-of-Thought~\cite{wei2022chain}) to produce persistent monitoring policies without relying solely on localization.
\item Real-world deployment(Fig.~\ref{fig:beauty_shot}) of our framework on an underwater robot, demonstrating feasibility of efficient exploration in marine environments for various object of interests.
\item We will open-source our synthetic environments, real-world dataset, and our code to facilitate further underwater robotics monitoring capabilities.  
\end{enumerate}

The rest of this paper is organized in the following manner. First, in Section~\ref{sec:related_work}, we discuss the recent advances in the field. The problem formulation and proposed approach are presented in Section~\ref{sec:approach}. Experimental validation in simulation and the lab environment are discussed in Sections~\ref{sec:experiments}, ~\ref{sec:Result}, respectively. Finally, a conclusion with discussion on future work is presented in Section~\ref{sec:conclusion}.

\section{Related Work}
\label{sec:related_work}

Large Language Models have revolutionized the field of robotics as a source of commonsense reasoning that bridges multimodal input from the environment, such as textual instructions or visual observations, to high-level actions~\cite{pmlr-v205-ichter23a,huang2022language,liang2023code,mavrogiannis2023cook2ltl,huang2023instruct2act,brohan2023rt,mu2024embodiedgpt,walczak2025atlasv2}. These actions are often coupled with low-level controllers~\cite{liang2023code} or robot-specific APIs~\cite{mavrogiannis2025discovering} to yield explicit robot control commands and hence entire actionable motion plans. In our work, we leverage visual observations to generate high-level robot actions and bridge simulation to reality by generating control commands through an integrated PD controller. While our system yields high-level actions in simulation, our approach is modular and can be combined with complementary work that performs obstacle avoidance and local navigation in underwater environments~\cite{mane2024eroas,yang2025duvin}.

Underwater robot navigation has been studied extensively in the literature\cite{christensen2022recent,wu2019survey} with a rich line of work focusing on tackling localization challenges~\cite{JoshiIROS2019} posed by underwater environments~\cite{joshi20243d,xu2025never,joshi2025cascade,lin2024uivnav,XanthidisIROS2021,xanthidis2020navigation}.
Many approaches have been developed for underwater monitoring and exploration, with applications ranging from cave exploration~\cite{gupta2025demonstrating,chatzispyrou2025mapping} to oyster detection~\cite{lin2025odyssee, lin2023oysternet,karapetyan2025oysterbot,wang2024shellcollect}
However, they lack the idea of the feedback loop that our approach introduces to account for efficient exploration and replanning.

Recent advances include adapting LLMs and VLMs to underwater and marine robots, and researchers have even finetuned such models~\cite{bi2024oceangpt,nguyen2025llamarine,alawode2025aquaticclip} on ocean science tasks as they enable online decision making powered by commonsense reasoning, making them suitable for long-term autonomous monitoring without the constant need of a human in the loop. Xie et al.~\cite{xie2025never} present an LLM-enhanced reinforcement learning controller that jointly tunes rewards and control parameters, improving maneuvering under extreme sea conditions. SeafloorAI~\cite{nguyen2024seafloorai} expands data resources for geology-focused perception, strengthening visual grounding and instruction following for seafloor surveys. Word2Wave~\cite{chen2025word2wave} offers a language-driven mission interface that maps high-level goals to efficient subsea deployments. Furthermore, OceanChat~\cite{yang2023oceanchat} enables natural language piloting with dynamic replanning, AquaChat++~\cite{saad2025aquachat++} coordinates multi-ROV aquaculture inspection with fault-aware task planning, and Autonomous Vehicle Maneuvering~\cite{kim2025autonomous} integrates visual grounding with LLM-guided path planning for surface vehicles. While OceanPlan~\cite{yang2024oceanplan} translates natural language instructions into AUV trajectories for long-horizon tasks, it does not construct a spatial memory of the environment. In contrast, our approach integrates a dynamically built occupancy map with structured Chain-of-Thought reasoning, enabling persistent and efficient long-term monitoring for benthic environments. These works collectively progress from instruction understanding to robust low-level control utilizing data and tooling that will allow long-horizon planning and safety-aware autonomy in the ocean. Our approach advances this line of work by coupling visual observations with planning and integrated control.

\section{Approach}
\label{sec:approach}

As depicted in Fig.~\ref{fig:overview}, our framework has three modules: perception, cognitive-aware
planning, and control. In the perception module, the ROV’s front-facing camera provides depth, semantic-segmentation, and RGB images. We fuse the depth and segmentation images to build an occupancy map, which is incrementally updated throughout the exploration process. In the planning module, the VLM receives these inputs together with a well-crafted prompt and chain-of-thought instructions; it focuses on the objects of interest and outputs high-level actions. Because VLM inference is relatively slow, we only use it for high-level navigation rather than direct low-level control. Finally, the control module executes those actions—translating discrete speed, direction, and turn-angle commands into motion to achieve efficient, long-term monitoring.

\subsection{Perception phase}
The perception phase encompasses how the robot senses, processes, and interprets information from its environment to build a structured representation that supports downstream decision-making and planning. At each step, the ROV collects front-facing RGB, depth, and semantic segmentation images. These multimodal inputs are fused through a multi-stage pipeline to generate spatial representations in the form of occupancy maps. Depth pixels are first transformed into 3D point clouds and projected into global coordinates using camera-to-world transformations. Given a pixel \((u,v)\) with depth \(z(u,v)\) and intrinsics \((f_x,f_y,c_x,c_y)\), its 3D position in the world frame is obtained by
\begin{equation}
\mathbf{p}_w(u,v) \;=\; 
\Pi\!\left(
\mathbf{T}_{c}^{w}
\begin{bmatrix}
\frac{(u-c_x)}{f_x}\,z(u,v) \\[6pt]
\frac{(v-c_y)}{f_y}\,z(u,v) \\[6pt]
z(u,v) \\[6pt]
1
\end{bmatrix}
\right),
\label{eq:backproj_world}
\end{equation}
where \(\Pi([x\; y\; z\; w]^\top) = [x/w,\; y/w,\; z/w]^\top\), subject to the depth validity constraint
\begin{equation}
z_{\min} \le z(u,v) \le z_{\max}.
\label{eq:depthvalid}
\end{equation}
where 
\begin{itemize}
    \item \(z_{\min}\) is the minimum reliable depth (e.g., \(0.1\,\text{m}\)) used to suppress sensor noise or self-reflections, and
    \item \(z_{\max}\) is the maximum reliable depth (e.g., \(20\,\text{m}\)) beyond which measurements are treated as invalid due to noise and range limitations.
\end{itemize}

Height thresholds classify the resulting points: elevated regions are marked as obstacles, while target objects (e.g., oysters) are filtered using semantic labels. 
Let \(m(u,v)\in\{0,1\}\) denote the semantic mask for the object of interest (e.g., oysters), with \(m=1\) indicating a positive detection. After applying the depth-validity constraint and the camera-to-world transform (Equation~\ref{eq:backproj_world}), we define the set of valid world points
\begin{equation}
\mathcal{P}_{\mathrm{valid}}=\left\{\mathbf{p}_w=(x_w,y_w,z_w)^\top \;:\; z_{\min}\le z(u,v)\le z_{\max}\right\}.
\label{eq:Pvalid}
\end{equation}
We partition \(\mathcal{P}_{\mathrm{valid}}\) into three disjoint regions:
\begin{align}
\mathcal{P}_{\text{obj}} &= \left\{\mathbf{p}_w\in\mathcal{P}_{\mathrm{valid}} \;:\; m(u,v)=1\right\}, 
\label{eq:Pobj}\\[4pt]
\mathcal{P}_{\text{obs}} &= \left\{\mathbf{p}_w\in\mathcal{P}_{\mathrm{valid}} \;:\; h_{\min}\le z_w \le h_{\max}\right\}\setminus \mathcal{P}_{\text{obj}}, 
\label{eq:Pobs}\\[4pt]
\mathcal{P}_{\emptyset} &= \mathcal{P}_{\mathrm{valid}} \setminus \bigl(\mathcal{P}_{\text{obj}}\cup \mathcal{P}_{\text{obs}}\bigr),
\label{eq:Pfree}
\end{align}
so that \(\mathcal{P}_{\mathrm{valid}}=\mathcal{P}_{\text{obj}}\;\dot{\cup}\;\mathcal{P}_{\text{obs}}\;\dot{\cup}\;\mathcal{P}_{\emptyset}\).
Here, \(\mathcal{P}_{\text{obj}}\) denotes target objects on (or near) the seafloor using semantics, \(\mathcal{P}_{\text{obs}}\) represents the elevated structures considered as obstacles, and \(\mathcal{P}_{\emptyset}\) defines empty (free) space.

Given the robot’s grid pose at time \(t\), denoted by \(\mathbf{c}_t\) with yaw \(\psi_t\), the occupancy map is updated using a ray-casting process within the camera’s field-of-view sector \((\psi_t,\theta)\) and sensing range \(d_{\max}\) over the free space \(N\). This yields the visible set \(V_t\), which incrementally updates the explored mask, \(E_t\):
\begin{align}
E_t &= E_{t-1}\ \cup\ V_t, \label{eq:fog_update}\\[6pt]
V_t &= \mathrm{RayCast}\!\left(\mathbf{c}_t,\psi_t,\theta,d_{\max},N\right).
\label{eq:fog_cast}
 \end{align}

\begin{figure}
    \centering
    \includegraphics[width=0.5\linewidth]{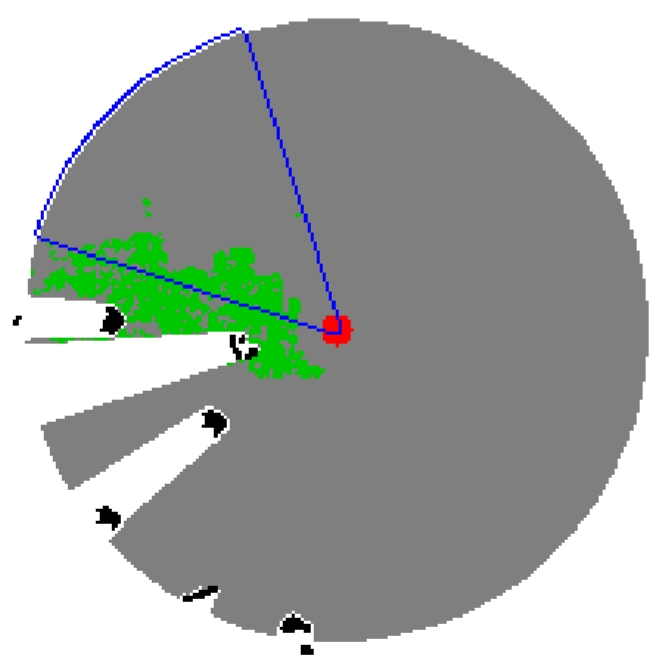}
    \caption{FOV demonstration on the occupancy map}
    \vspace{-5mm}
    \label{fig:occmap}
\end{figure}
As illustrated in Fig.~\ref{fig:occmap}, the blue cone represents the instantaneous field of view (FOV) of the camera and the occupancy map is generated by tracing rays through this FOV. Once a ray intersects an occupied cell, all subsequent cells along that line of sight are marked as \emph{unknown}, since the sensor cannot see beyond obstacles. These occluded regions appear as white gaps in the occupancy map and can only be revealed through exploration from different viewpoints. This distinction between navigable space, occupied space, target objects, and unknown space is critical for safe navigation and for guiding next-best-view exploration. The occupancy grid also incorporates safety margins to account for the robot’s physical dimensions. In real-world trials, the same RGB, depth, and segmentation modalities are captured by the ROV’s onboard sensors and the occupancy map is incrementally updated using depth, semantic segmentation, and pose estimates. This map provides the VLM with both a concise representation of the surrounding structure and a persistent memory of explored regions.

The exploration process proceeds in two stages. First, the robot performs an in-place rotation to acquire a panoramic view, constructing a 360° occupancy map of its immediate surroundings. This initialization phase enables the robot to identify safe entry directions and plan its first movements. In the second stage, the robot executes exploration actions guided by next-best-view predictions, incrementally expanding its map and knowledge of the environment.



\subsection{Cognitive-Aware Planning phase}
After the cognition phase, the VLM receives the front-camera RGB, segmentation, and depth images, along with the occupancy map. These inputs provide the ROV with situational context, the locations of target objects, and their spatial distribution. To ensure the model understands the task, we supply a carefully crafted prompt that enumerates the inputs and units, defines the mission, specifies clear standards for abstract instructions, outlines exploration guidance and decision priorities, and lists the available actions. Since we want to have long-term monitoring on our target objects, we define our task:  ``\textit{Your mission is to efficiently and comprehensively discover and map all oyster clusters (target object) on the seafloor.}" Upon experimentation, we find that VLMs can struggle with vague directives (e.g., “\textit{finish exploring the nearby oyster area}”), therefore we define completion precisely: green (oyster) regions on the occupancy map must be fully enclosed by gray (explored) cells, with no white (missed) patches inside. As with human operators, explicit guidance is crucial; without it, the robot tends to move randomly. We therefore instruct the agent to follow the local continuity of oyster distributions, expanding coverage around nearby clusters rather than jumping to distant ones, to improve efficiency. Nevertheless, a well-crafted prompt alone does not always yield coherent inference; the model can still lack a clear reasoning path.

Thus, we leverage Chain of Thought (CoT)~\cite{wei2022chain}, a sequence of intermediate reasoning steps that can substantially improve complex reasoning. We design a hand-crafted CoT prompt that encodes an efficient, human-like search logic and yields a robust, transparent reasoning path that consists of the following steps:
\begin{enumerate}
    \item Distribution analysis: infer local oyster layout and density from segmentation/occupancy. 
    \item Select current target area: choose the nearest/high-value frontier on the oyster boundary.
    \item Completion check: verify whether the current region satisfies the completion criterion.
    \item Select next target/frontier: prioritize adjacent, denser oyster areas; avoid long, low-yield jumps.
    \item Safety \& feasibility: confirm standoff distance and collision-free motion using depth/occupancy.
    \item Action selection: output direction (left/right/forward), turn angle, and step length from the discrete action set.
\end{enumerate}

\subsection{Control}

We use a planar PD controller for positioning and yaw regulation, while vertical motion is stabilized by the vehicle’s depth/altitude hold. Let $p = [x, y, z]^\top$ be the position of the robot and $\psi$ be its yaw angle. Like most ROVs, the robot is dynamically stable in its roll and pitch. For a desired waypoint $p^*$ with desired heading $\psi^*$, the body-frame pose error by rotating the world frame error into the vehicle frame $R(\tilde{\psi})$, $e_p$ is defined as 
\begin{equation}
    e_p = R(\tilde{\psi})^\top \tilde{p}, \quad R(\hat{\psi)} = \begin{bmatrix}
        \cos \hat{\psi} & -\sin \hat{\psi}\\
        \sin \hat{\psi} & \cos \hat{\psi}
    \end{bmatrix}
\end{equation}
where $\tilde{(\cdot)}$ denotes the error $(\cdot)^* - \hat{(\cdot)}$. The velocity command $[u_c, v_c]^\top$ and angular yaw rate $r_c$ sent to the robot is driven by the PD controller
\begin{align}
    \begin{bmatrix}
        u_c \\
        v_c
    \end{bmatrix} &= K_p e_p - K_d \hat{v}_b \\
    r_c &= k_\psi \tilde{\psi} - k_r \hat{r}
\end{align}
where $\hat{v}_b = [\hat{u}, \hat{v}]^\top$ and $\hat{r}$ are the estimated surge/sway velocities and yaw rates, respectively.

An IMU–DVL estimator supplies this estimated state feedback for the PD law. We use an invariant extended Kalman filter (InEKF) \cite{joshi20243d, potokar2021invariant, barrau2016invariant} as well as a Cascade Iteratively Preconditioned Gradient Descent (C-IPG) \cite{joshi2025cascade} observer to propagate the discrete-time inertial dynamics and update the state with body-frame water-relative velocities from the DVL (and pressure depth) when available. These methods give us robust attitude/velocity estimates even during aggressive yawing and intermittent DVL bottom-lock. The update treats DVL as a direct velocity observation in the vehicle frame and applies zero-velocity or near-hover pseudo-measurements whenever the policy asks the robot to “Stop,” which helps bound drift.

Together, these layers form a simple stack: language actions → short setpoint segments → PD tracking, with state supplied by an observer in the form of InEKF or C-IPG (IMU+DVL) and gently regularized by continuous-time alignment when the scene supports it. We explicitly decoupled perception (oyster segmentation) and control stability, thereby maintaining safety.
\section{Experimental Setup}
\label{sec:experiments}

\subsection{Environment}
\begin{figure*}[ht!]
\vspace{3mm}
\includegraphics[width=0.95\textwidth]{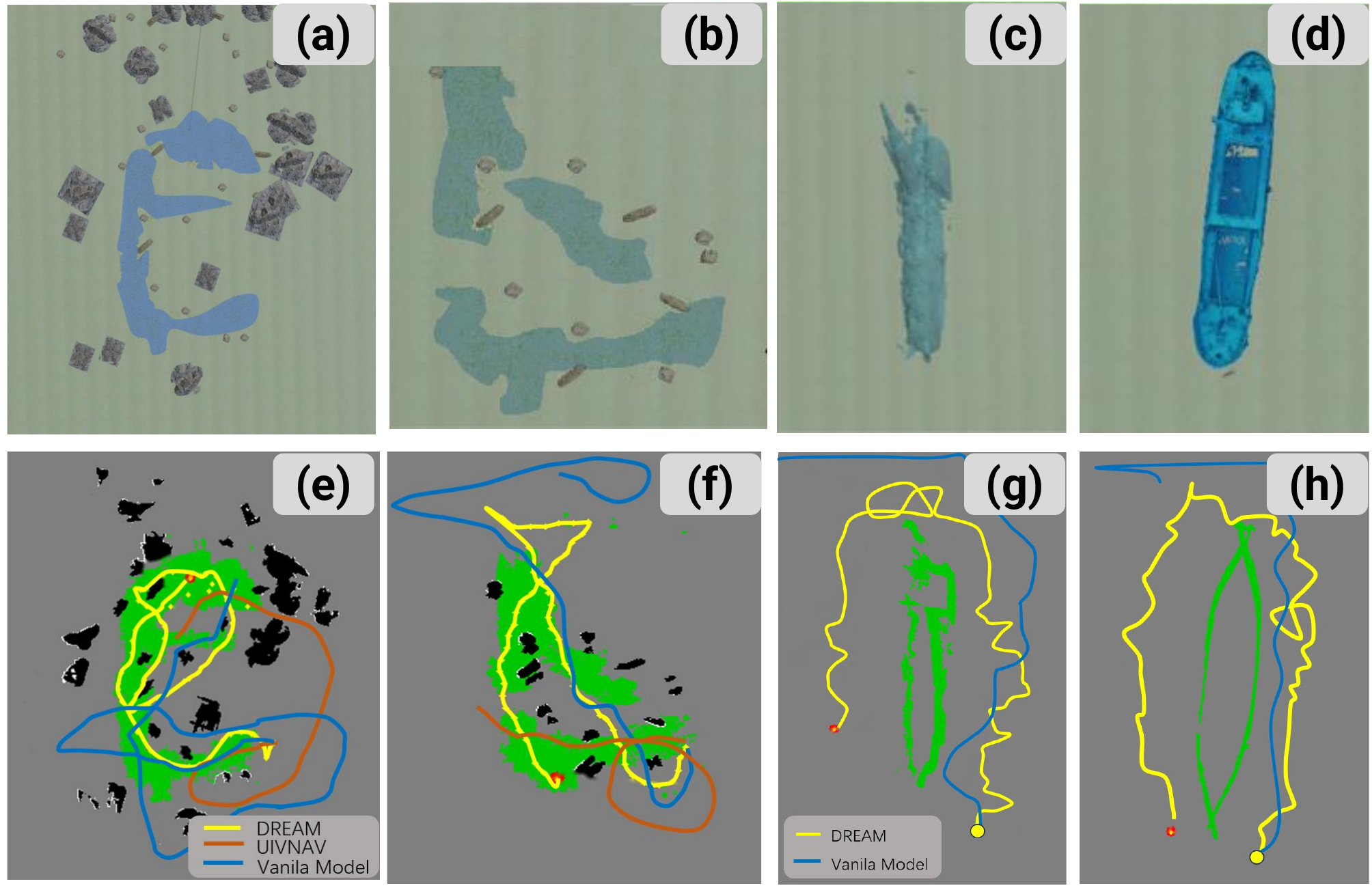}
\centering
\caption{Testing environments and algorithmic comparisons. Top row: simulated oyster reef (left) and shipwreck (right) environments. Bottom row: comparison of our framework against baseline methods, showing improved efficiency and coverage in both oyster and shipwreck monitoring tasks.}
\vspace{-5mm}
\label{fig:comparison}
\end{figure*}

We evaluate our approach in Oystersim \cite{lin2022oystersim} across 15 environments with obstacles, of which $10$ are oyster reefs and $5$ are shipwrecks (Fig.~\ref{fig:comparison} top row). Our environment's length range and width range are both from $40$ meters to $50$ meters. To mimic natural oyster morphology and show our framework's adaptability, we put fringing, string (linear), and patch reefs, appearing either individually or in combination. For the shipwreck set, we select five distinct wreck models and place them on the seafloor. 

\subsection{Metrics}
For all environments, we report the total exploration time, coverage percentage (for oysters we consider the number of oysters and for shipwrecks the area). The total exploration time scores the efficiency of an agent by measuring the total time it takes to finish the entire exploration. An agent’s efficiency is inversely related to its exploration time; greater time denotes lower efficiency and vice versa. We set max steps to $200$. If in $200$ steps exploration has not concluded, we set the  total exploration time to $200$ steps. The percentage of coverage shows the quality of exploration. If equal to $100$, it shows that the agent can cover all target objects in the desired area. 
\subsection{Baseline}
We compare our framework with UIVNAV~\cite{lin2024uivnav}, a previous AI-based method for zero-shot object navigation. UIVNAV is an end-to-end underwater navigation approach that leverages intermediate representations to enable zero-shot transfer to new target objects without retraining.
Beyond method comparisons, we evaluate our crafted chain-of-thought (CoT) against a naïve prompt (no CoT) to isolate the benefit of structured reasoning. Both prompting strategies use the same VLM; in our experiments, we adopt GPT-5\cite{openai2025chatgpt5} as the reasoning model.

\section{Results}
\label{sec:Result}
\subsection{Comparison on oysterbed}
As shown in Table~\ref{tab:oyster_comparison} and Fig.~\ref{fig:comparison}(e)(f), our framework outperforms the baseline and the vanila model. In this figure, guided by our framework (yellow trajectory), the ROV follows the oyster distribution, remains focused on the target regions, and exhibits minimal redundant exploration. In contrast, the baseline and vanila model do not track the oyster pattern and often diverge to unrelated areas. More specifically, in oyster scenarios, our framework uses $26\%$ less time than UIVNAV and achieves $98.26\%$ coverage within the same time budget. In contrast, the baseline only has a $55.46\%$ average coverage within the limit, and the vanila model has an average coverage of $90.59\%$. Therefore, we observe that no baseline can cover more oysters than our approach. On the other hand, our framework efficiently explores and maps the entire wreck without collisions, attaining an average coverage of $100\%$ with an average exploration time of $145$ steps.

\subsection{Comparison on shipwreck}
Using vanila mode, the ROV exhibits inconsistent behavior. As shown in Table~\ref{tab:Vlm_comparison} and Fig.~\ref{fig:comparison}(g)(h), our path (yellow trajectory) circumnavigates the wreck and achieves full coverage without collisions, whereas the vanilla model fails to cover the entire structure and at times drifts outside the boundary. Finally, the vanilla model has $200$ steps with $60.23\%$ coverage in shipwreck environments. Coupled with our framework, the ROV uses $27.5\%$ fewer steps and achieves $66.03\%$ higher coverage in shipwreck scenarios.

\begin{table}[t]
  \centering
  \vspace{3mm}
  \caption{Comparison with UIVNAV (Oyster)} 
  \label{tab:oyster_comparison}
  \begin{tabular}{|c|c|c|}
    \hline
    \diagbox{Method}{Prompt Type} & Avg. Exploration &Avg. Coverage \\ & Time (steps) & Rate (\%) \\ \hline
    UIVNAV &200 &55.46\% \\ \hline
    Our Framework &137.3 &98.26\% \\ \hline
  \end{tabular}
\end{table}

\begin{table}[t]
  \centering
  \caption{Comparison with original VLM} 
  \label{tab:Vlm_comparison}
  \setlength{\tabcolsep}{5pt} 
  \begin{tabular}{|c|c|c|}
    \hline
    \diagbox{Environment}{Framework} & Our framework & GPT-5 \\ \hline
    \shortstack{Exploration Time-Oyster \\ (steps)} & 137.3 & 179.7 \\ \hline
    \shortstack{Exploration Time-Shipwreck \\ (steps)} & 145 & 200 \\ \hline
    \shortstack{Average Coverage Rate- \\ Oyster} & 98.26\% & 90.59\% \\ \hline
    \shortstack{Average Coverage Rate- \\ Shipwreck} & 100\% & 60.23\% \\ \hline
  \end{tabular}
  \vspace{-3mm}
\end{table}

\subsection{Real-world deployment}

To demonstrate feasibility in real-world settings, the framework was tested in a scenario where oyster shells and a pipe (to simulate a shipwreck) were laid out in a tank, and a BlueROV2 robot platform was used to survey the environment. The robot platform used is a BlueROV2 equipped with an onboard camera. Other sensors on the BlueROV2 include an onboard IMU and a retrofitted Waterlinked DVL-A50 to localize and control the ROV's position.

\begin{figure}[t!]
    \centering
    \vspace{3mm}
    \includegraphics[width=0.4\textwidth]{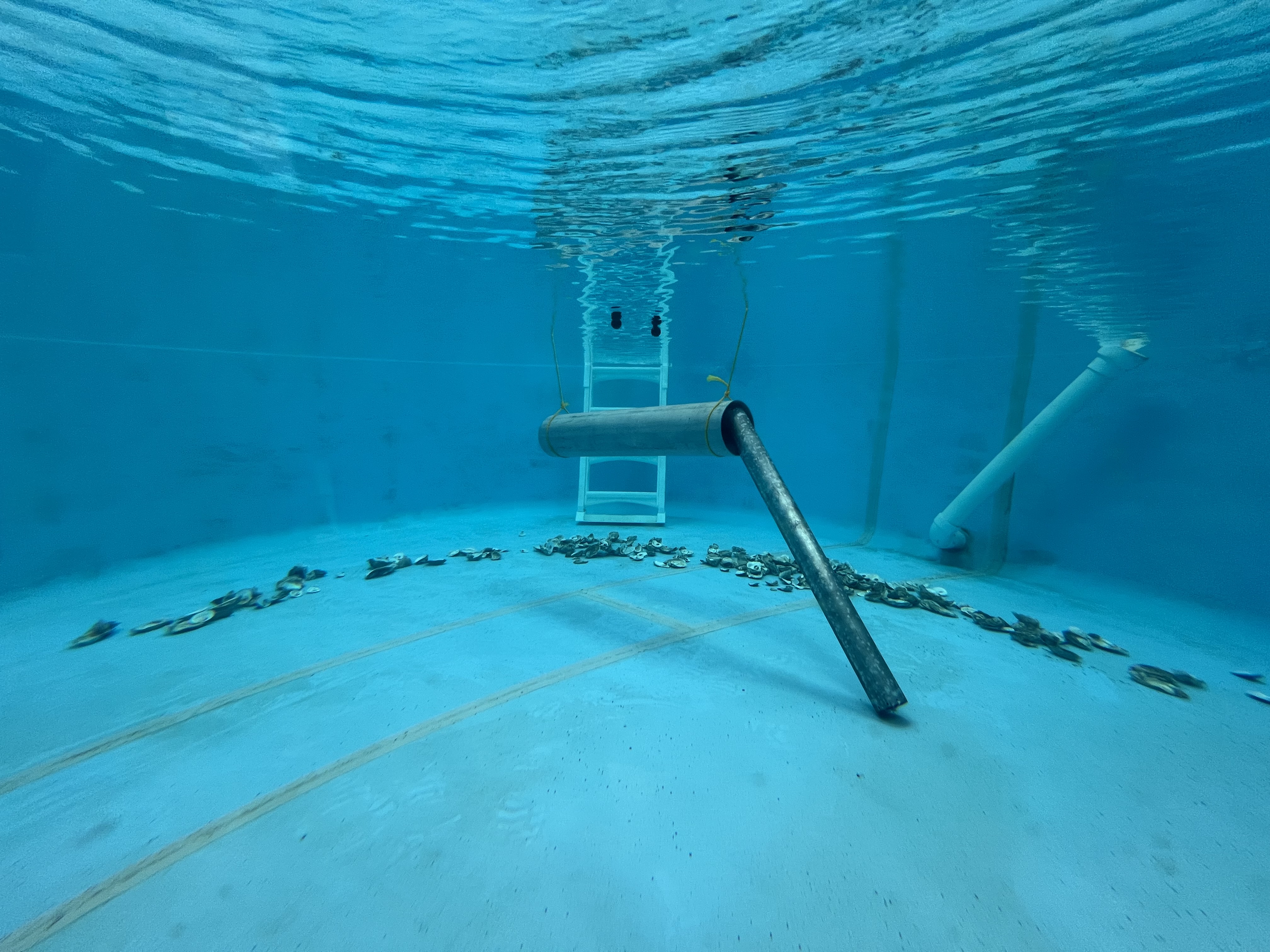}
    \caption{Pool setup with oyster shells arranged in a circular arc and two pipes positioned to emulate a shipwreck.}
    \label{fig:exp_setup}
\end{figure}

We evaluated the system in a circular tank that is $12ft$ in diameter and $5ft$ deep. $200$ oyster shells were arranged in a circular arc pattern and two pipes linearly as shown in Fig. \ref{fig:exp_setup}. During trials, the vehicle faced realistic pool-scale disturbances: slow cross-currents, tether drag, small turn-radius asymmetries due to hydrodynamic coupling, and intermittent visual degradation from surface reflections. Despite these effects, the pipeline executed end-to-end in real time, with the front camera producing frames for on-board oyster segmentation, and the PD controller ensuring stability and positions by estimating odometry from C-IPG, as well as InEKF. 

\begin{figure}[t!]
    \centering
    {\includegraphics[width=\columnwidth]{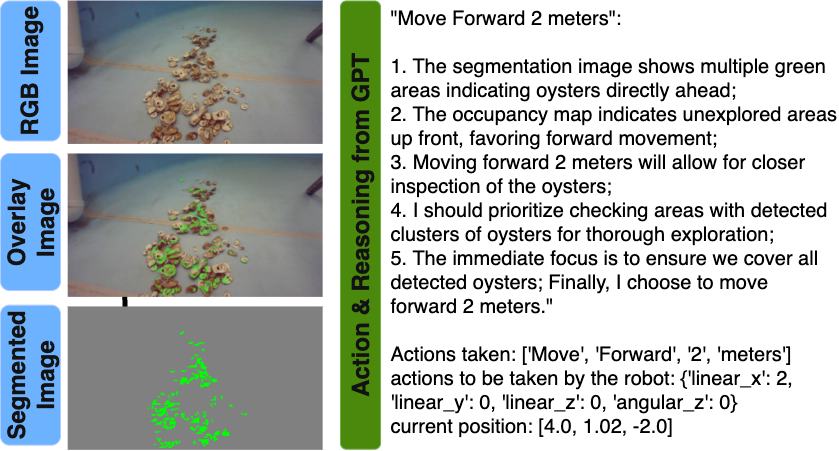}}
    \caption{Example of a real-world experiment and the low-level actions produced by our framework.}
    \label{fig:pool_experiment}
\end{figure}

A representative run is shown in Fig.~\ref{fig:pool_experiment}. Starting without a prior map, the robot first stabilized yaw and altitude, then advanced with gentle, $0.1m$ steps, favoring directions where segmented oysters densified. The step size of $0.1\text{m}$ was selected based on the limited dimensions of the tank of $15ft$. Qualitatively, this experiment exhibits near-complete traversal of the oyster patch with only a few missed interior pockets.





  

\section{Conclusion and Future Work}

\label{sec:conclusion}
In this paper, we propose a VLM-based framework for underwater monitoring. The system constructs an occupancy map to provide the VLM with a compact overview of the surroundings and a persistent memory of previously visited locations. Conditioned on target object, the VLM outputs high-level actions. By supplying a well-crafted prompt and a chain-of-thought (CoT) scaffold, we endow the model with stronger situational understanding and efficient, human-like navigation capabilities. In comprehensive simulation and real-world experiments, our framework demonstrates superior performance, adaptability, efficiency, and complete coverage capability. Although our experiments are conducted in relatively small-scale environments due to time and hardware limitations, they serve as an initial step towards persistent long-term monitoring.


For future work, we plan to leverage DINOv3~\cite{simeoni2025dinov3} features as input to improve generalizability, expand the map size and environmental complexity to approximate long-term deployments, and increase the number of planning steps. Furthermore, we intend to train a compact local VLA model based on Mamba\cite{aalishah2025edgenavmamba} and apply pruning\cite{kallakuri2025magrip,kallakurik2025enabling} and extreme quantization\cite{walczak2025bitmedvit} to enable efficient on-device inference, along with more rigorous evaluations to assess system robustness. Last but no least, we aim to extend our framework to 3D exploration in cave-like environments, where irregular structures and constrained visibility present additional challenges for autonomous navigation.

\bibliographystyle{IEEEtran}
\bibliography{references}

\end{document}